\documentclass[fleqn,10pt,twocolumn]{SICE20}

\usepackage{graphicx}
\usepackage{amsmath}
\usepackage{commath}
\usepackage{subfigure}

\title{Towards Spiral Brick Column Building Robots}

\author{Yasser Ashraf ${}^{1}$, Ahmed Abdallah${}^{1}$, Abdelhaleem Osman${}^{1}$, Victor Parque${}^{2}$ and Samy Assal${}^{1}$}

\speaker{This paper has appeared in the Proceedings of the SICE Annual Conference 2020, pp. 1448 - 1451.}

\affils{ ${}^{1}$Department of Mechatronics and Robotics, Egypt-Japan University of Science and Technology, Alexandria, Egypt\\
(E-mail: yasser.attia@ejust.edu.eg, ahmed.abdelgayed@ejust.edu.eg, abdelhaleem.saad@ejust.edu.eg, samy.assal@ejust.edu.eg)\\
${}^{2}$ Department of Modern Mechanical Engineering, Waseda University, Tokyo, Japan\\
(E-mail: parque@aoni.waseda.jp)\\
}

\abstract{Automation in construction has the potential to expand the technological landscape of labor intensive tasks, and bring gains in efficiency and productivity to sustain global competitiveness. In this paper we propose a task-level approach for assembly of spiral brick columns. Our extensive computational simulations using the generalized models of spiral brick columns show the feasibility, the effectiveness and efficiency of our proposed approach. Our results offer the potential to use robots in automated construction of spiral brick columns with utmost efficiency.
}

\begin{document}

\maketitle

\section{Introduction}

\begin{figure*}[h]
\begin{center}
\includegraphics[width=0.8\textwidth]{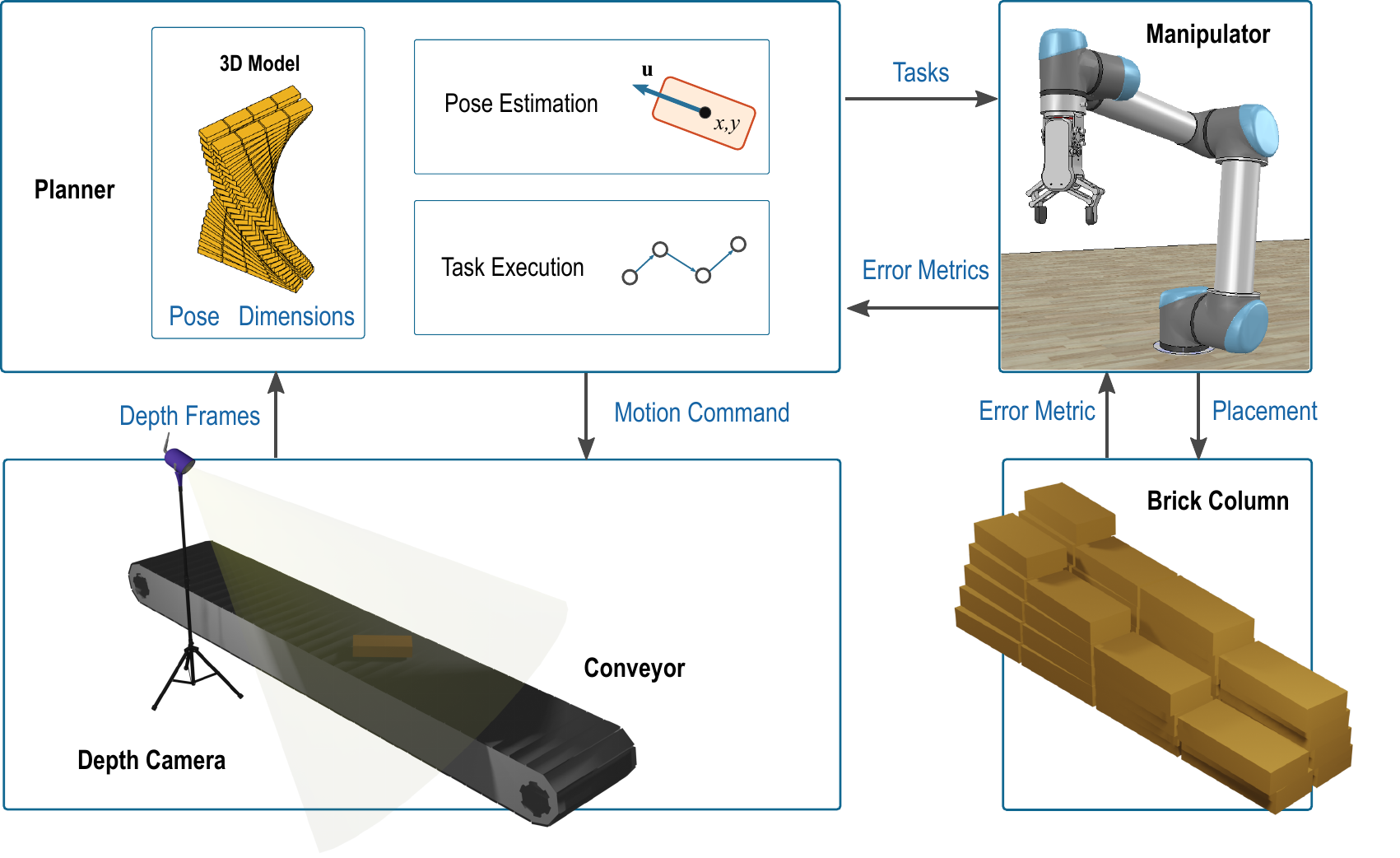}
\caption{Basic idea of he general framework in our approach. The manipulator is attached to a gripper and is tasked with assembling a user-defined spiral brick column by using incoming bricks from a belt conveyor.}
\label{gen}
\end{center}
\end{figure*}

The autonomous assembly of bricks in construction buildings consists in locating, picking, transporting and assembling brick objects to build pre-defined structures. Since the manual teaching of such tasks is inefficient, the model-driven task planning is potential to expand productivity frontiers.

Past research has explored the assembly of wall scenarios in construction by serial manipulators\cite{bock96,helm12,kriz20}. Also, mobile robots have been used to evaluate the assembly task\cite{lussi18}, to aid in localization in construction sites\cite{inoue11}: to automate the generation of robot programs for prefabrication of walls\cite{ben18} and to integrate with building information modeling (BIM) to develop plans to construct walls, stairs, and a pyramid with up to 30 bricks\cite{ding20}

Nowadays, there already exists a number of commercially available initiatives for the assembly of buildings in the market. Developed by Construction Robotics, SAM 100 is the first commercially bricklaying robot for straight-line mason construction, and MULE 135 (Material Unit Lift Enhancer) is a lift assisted device for handling and placing heavy materials in construction tasks\cite{sam15}. Developed by Fast-brick Robotics, Hadrian X is an automated bricklaying system able to configure the first multi-room house structure from a 3D CAD model without human intervention\cite{pivac}.

The full automation of bricklaying in arbitrary building structures is still unfeasible and requires human supervision. Although the use of robotics in straight line walls has been proposed, the study of building spiral brick columns has received little attention. In this paper, we propose a task-level planning of assembly of spiral brick columns and evaluate its feasibility and efficiency by using a generalized modeling to render a relevant set of spiral brick models and simulation environments (as shown by Fig. \ref{gen}). Our study is potential to build robot plans being effective for automated construction of aesthetic spiral brick columns.





\section{Proposed Approach}


%
%


\subsection{Spiral Brick Models}

\begin{figure*}[h]
\begin{center}
\includegraphics[width=0.98\textwidth]{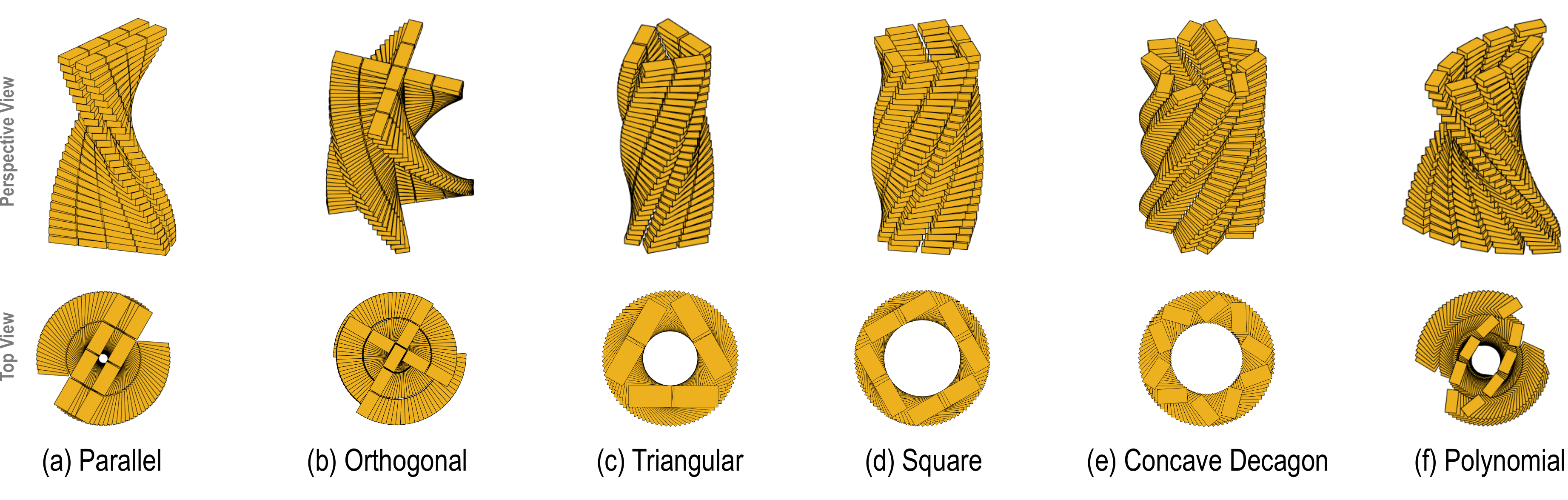}
\caption{Spiral brick column models considered in our study.}
\label{models}
\end{center}
\end{figure*}

We considered the modeling of spiral brick columns as shown by Fig. \ref{models}, which, according to the geometry of the base of the column and its configuration parameters, topologically distinct geometries are renderable. When the base of the spiral column is composed by segments with \emph{parallel} and \emph{orthogonal} configurations, we let the formation of $s$ segments each with $B_i$ block units. Then the margin $m$ between two blocks is computed by:

\begin{equation}\label{mreg}
m = B_i.l + \tau.w + \lambda(B_i - 1)
\end{equation}

\begin{equation}\label{tau}
\tau = \frac{1}{\tan(\frac{\theta}{2})}
\end{equation}
, where $B_i$ is the number of blocks in the $i$th segment of the base configuration of the spiral column, $l$ is the block length, $w$ is the block width, $\tau$ is the factor of the angle between segment configurations, $\theta$ is the angle between two segment configurations, and $\lambda$ is a user-defined coefficient for margin preference. Thus, to enable a spiral brick column we let $B_i = B_j$ for $i, ~j \in [s], i \neq j$, when the segments are in parallel configuration  ($\theta = \pi$), as portrayed by Fig. \ref{models}-(a), and $B_i \neq B_j$ for $i, ~j \in [s], i \neq j$, as portrayed by Fig. \ref{models}-(b), when the segments are in orthogonal configuration ($\theta = \pi/2$). The above-described configuration is a general mechanism which is a function of the number of segments and the configuration between segments. The segment base configuration is rotated around its centroid by an angle $\phi$ to form $L$ consecutive layers of spiral brick column.

When the base of the spiral column is a \emph{regular polygon}, the number of segments $s$ in the column base is equal to the number of edges $n$ of the regular polygon. Thus, the spiral brick column is renderable by the geometry of the base polygon (the angle $\theta$ is computable from the number of sides $n$ of the regular polygon), and the configurations in the number of blocks in each edge of the polygon base. To allow the formation of a spiral column, we let $B_i = B_j$, for $i, ~j \in [n], i \neq j$, and let rotate the base polygon around its centroid by an angle $\phi$ to form $L$ consecutive layers. Examples of spiral brick columns wit regular polygonal configuration are portrayed by Fig. \ref{models}(c)-(e) in which the base polygon consists of not only convex geometry (triangle and square) but also concave geometry (concave decagon) .


It is also possible to build spiral column configurations by using polynomial functions. Here, the pose of the bricks are allowed to follow the trajectory of an arbitrary polynomial function $f$ over a fixed domain and range. Here, the margin $m$ between two bricks is computed by:

\begin{equation}\label{}
   m = w. \sin \Big(\frac{\pi - \theta}{2} \Big) + \kappa
\end{equation}

, where $\theta$ is the angle between two successive bricks, and $\kappa$ is a user defined coefficient for margin preference. In order to allow the formation of a spiral column, we let the polynomial function $f$ and $-f$ to form a closed loop, and let it rotate around its centroid by an angle $\phi$ over the plane to form $L$ consecutive layers.


\begin{figure*}[t!]
	\begin{center}
		\subfigure[Depth Map]{\label{fig3:c}\includegraphics[width=0.5\columnwidth]{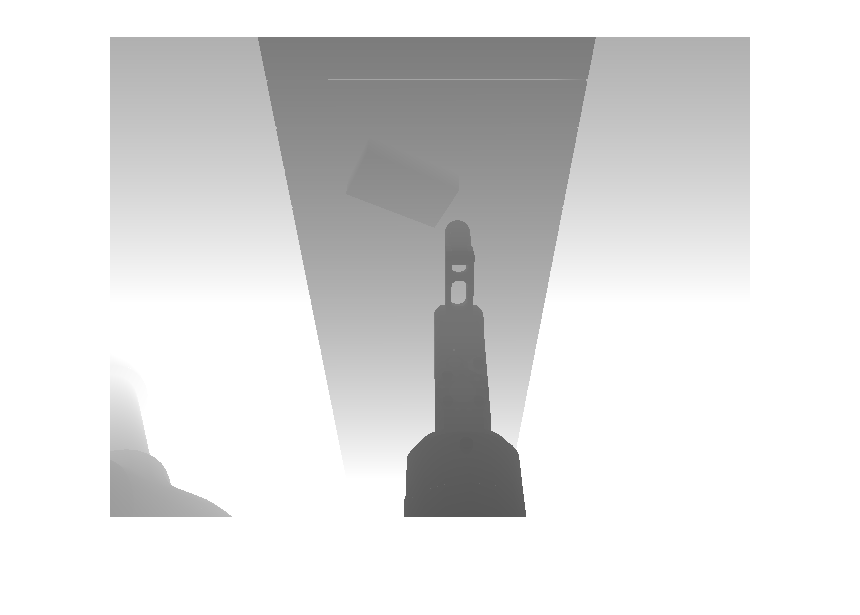}}
		\hfill
		\subfigure[Point Cloud]{\label{fig3:c}\includegraphics[width=0.5\columnwidth]{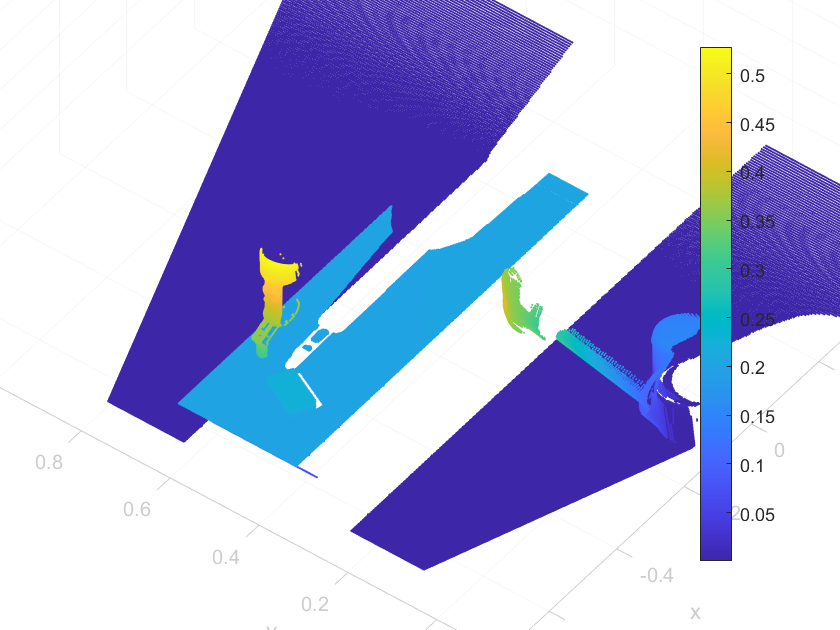}}
		\hfill
		\subfigure[Filtered Point Cloud]{\label{fig3:c}\includegraphics[width=0.5\columnwidth]{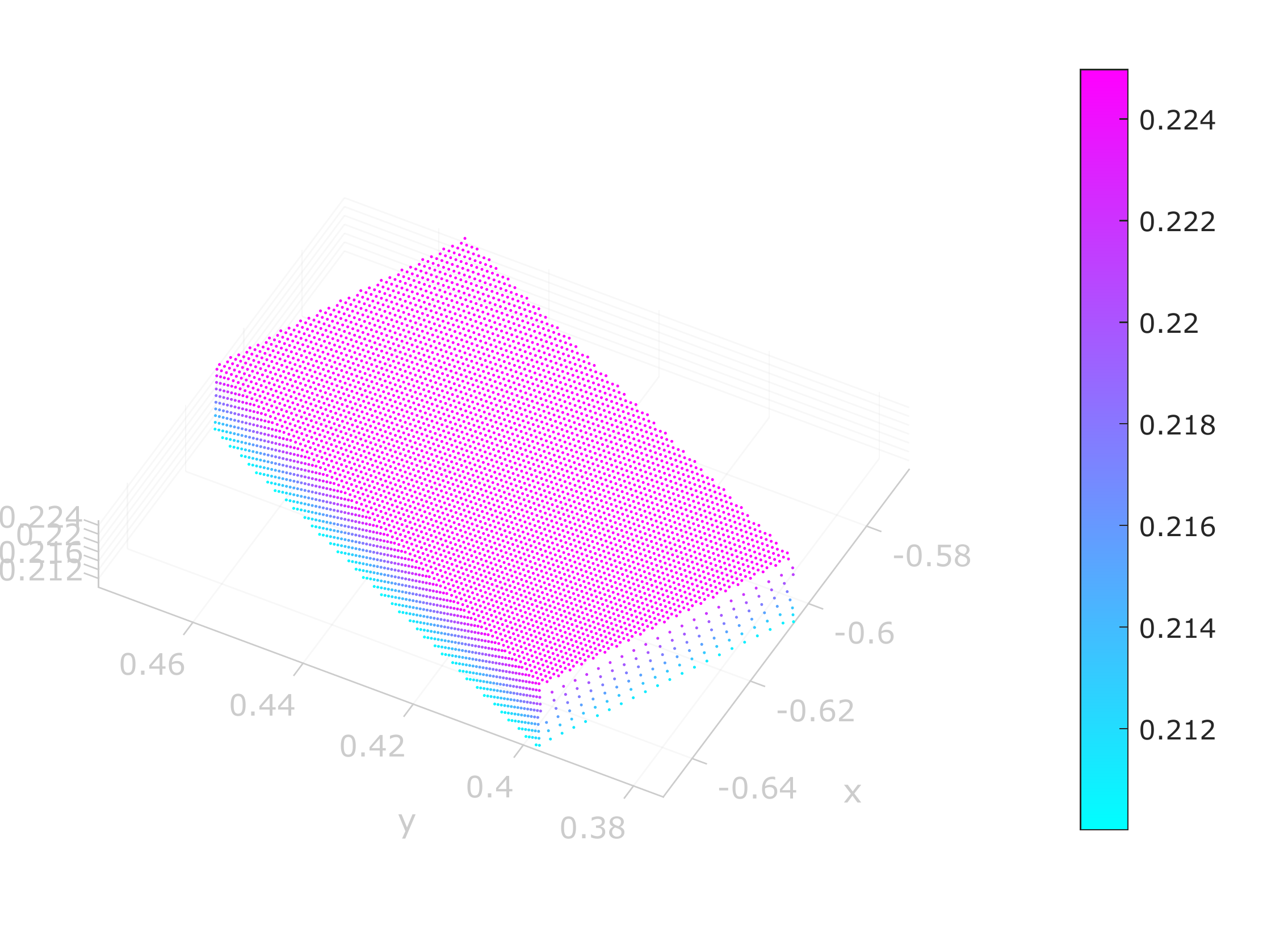}}
		\hfill
		\subfigure[Estimated Pose]{\label{fig3:c}\includegraphics[width=0.5\columnwidth]{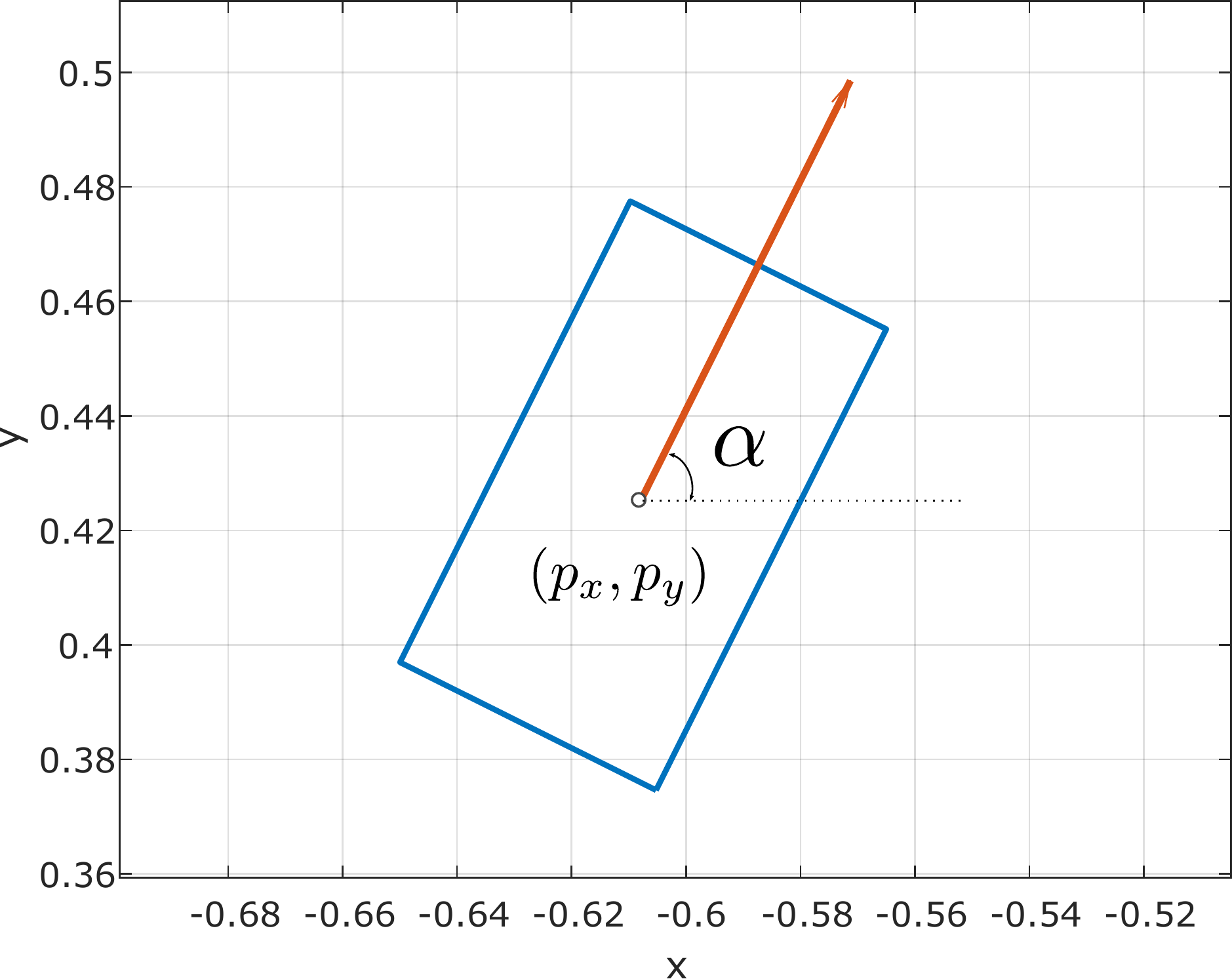}}
		\hfill
	\end{center}
	\caption{Pose estimation of arbitrary bricks on a conveyor.}
	\label{pcloud}
\end{figure*}

\begin{figure*}[t!]
	\begin{center}
		\subfigure[Position Error]{\label{perfoa}\includegraphics[width=0.98\columnwidth]{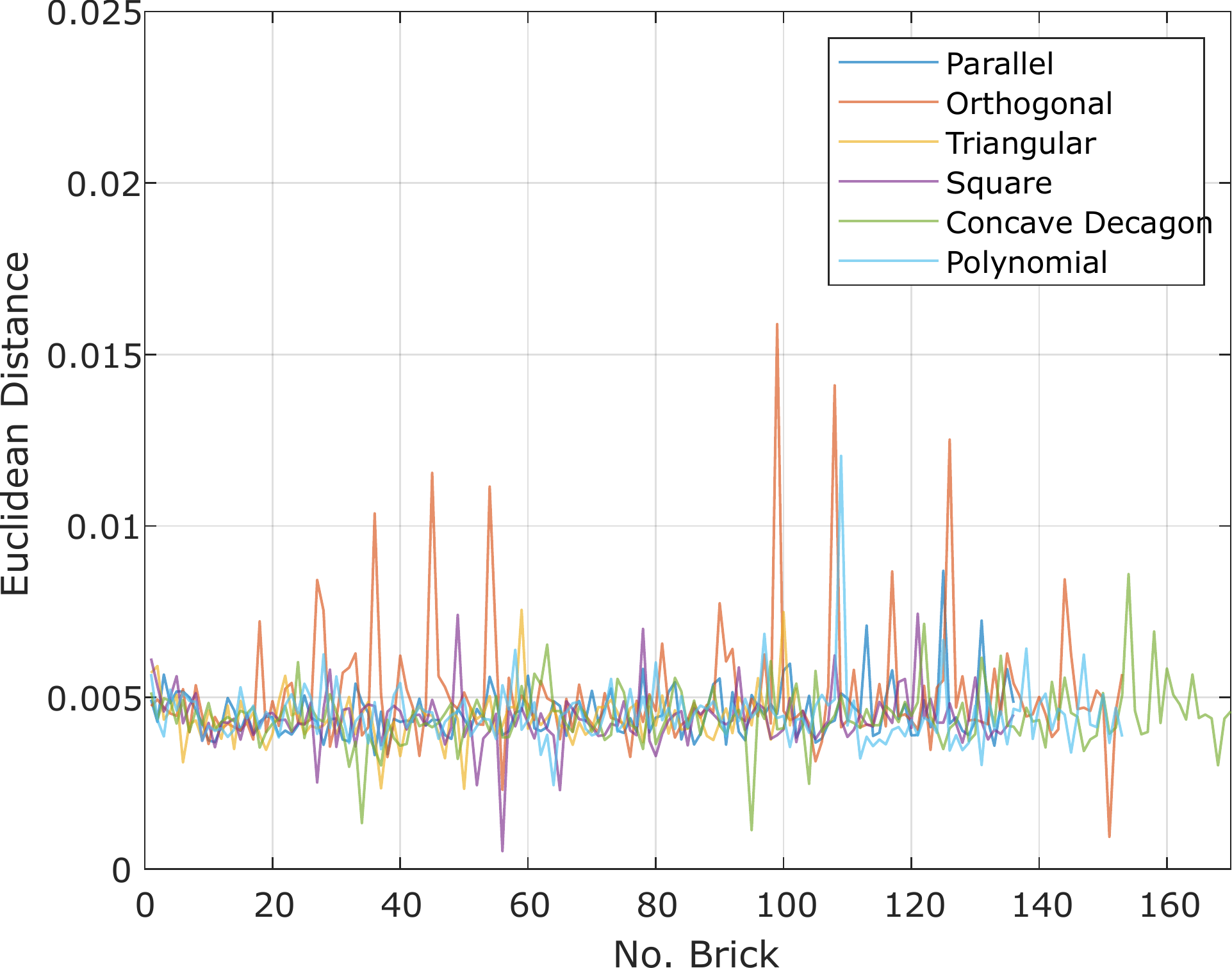}}
		\hfill
		\subfigure[Orientation Difference]{\label{perfob}\includegraphics[width=0.98\columnwidth]{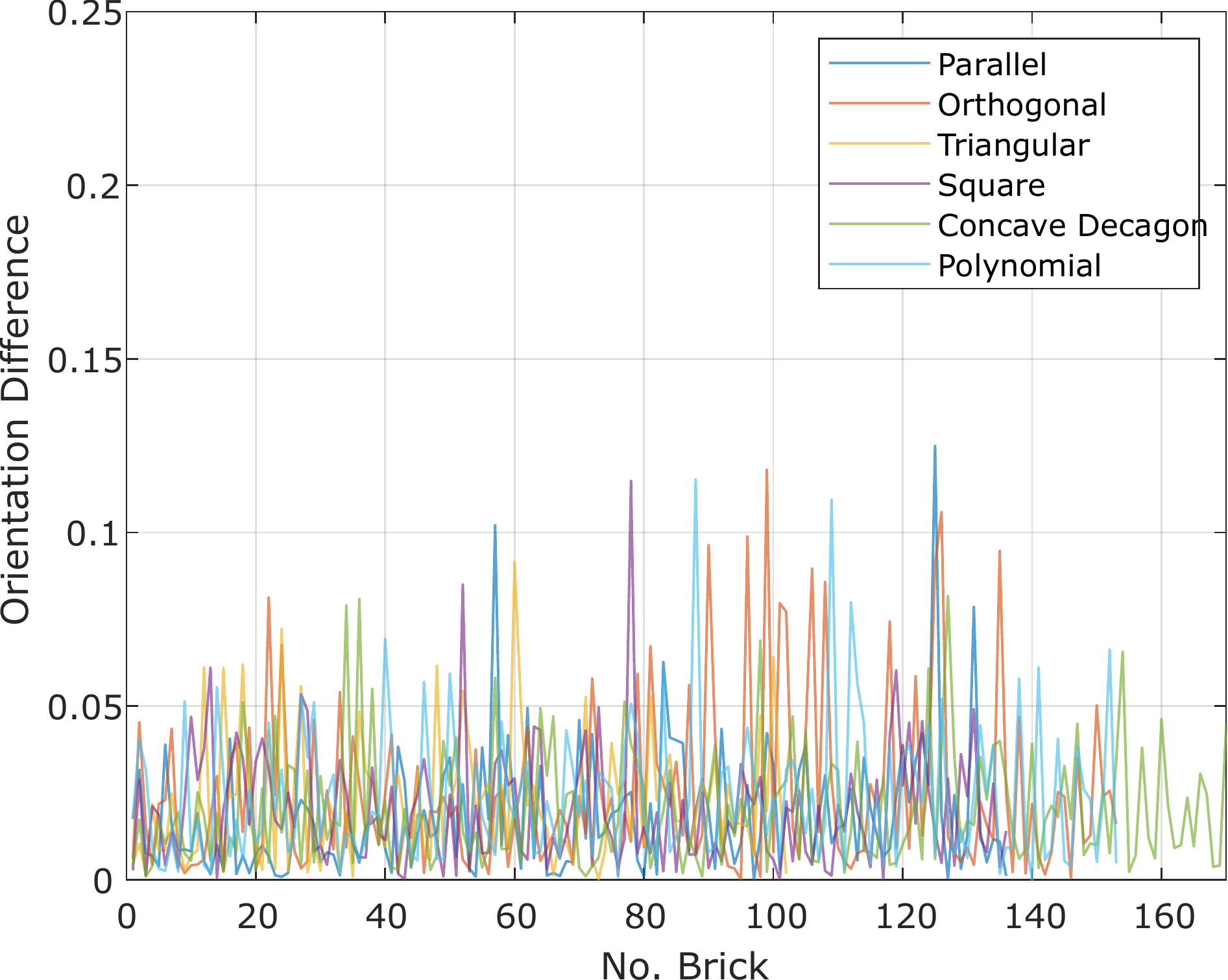}}
		\hfill
		\subfigure[Trajectory Time]{\label{fig3:c}\includegraphics[width=0.98\columnwidth]{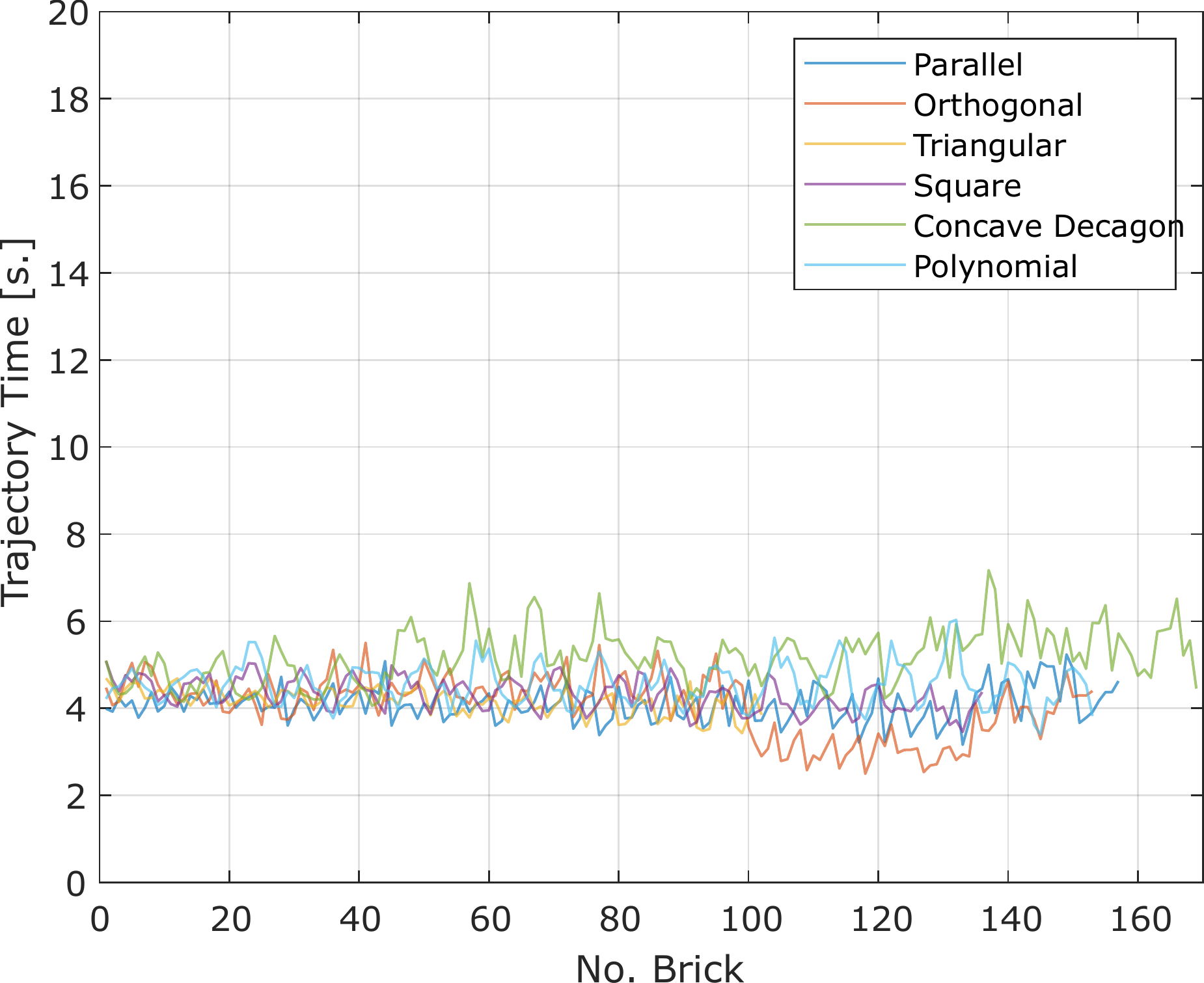}}
		\hfill
		\subfigure[Pose Estimation Time]{\label{fig3:c}\includegraphics[width=0.98\columnwidth]{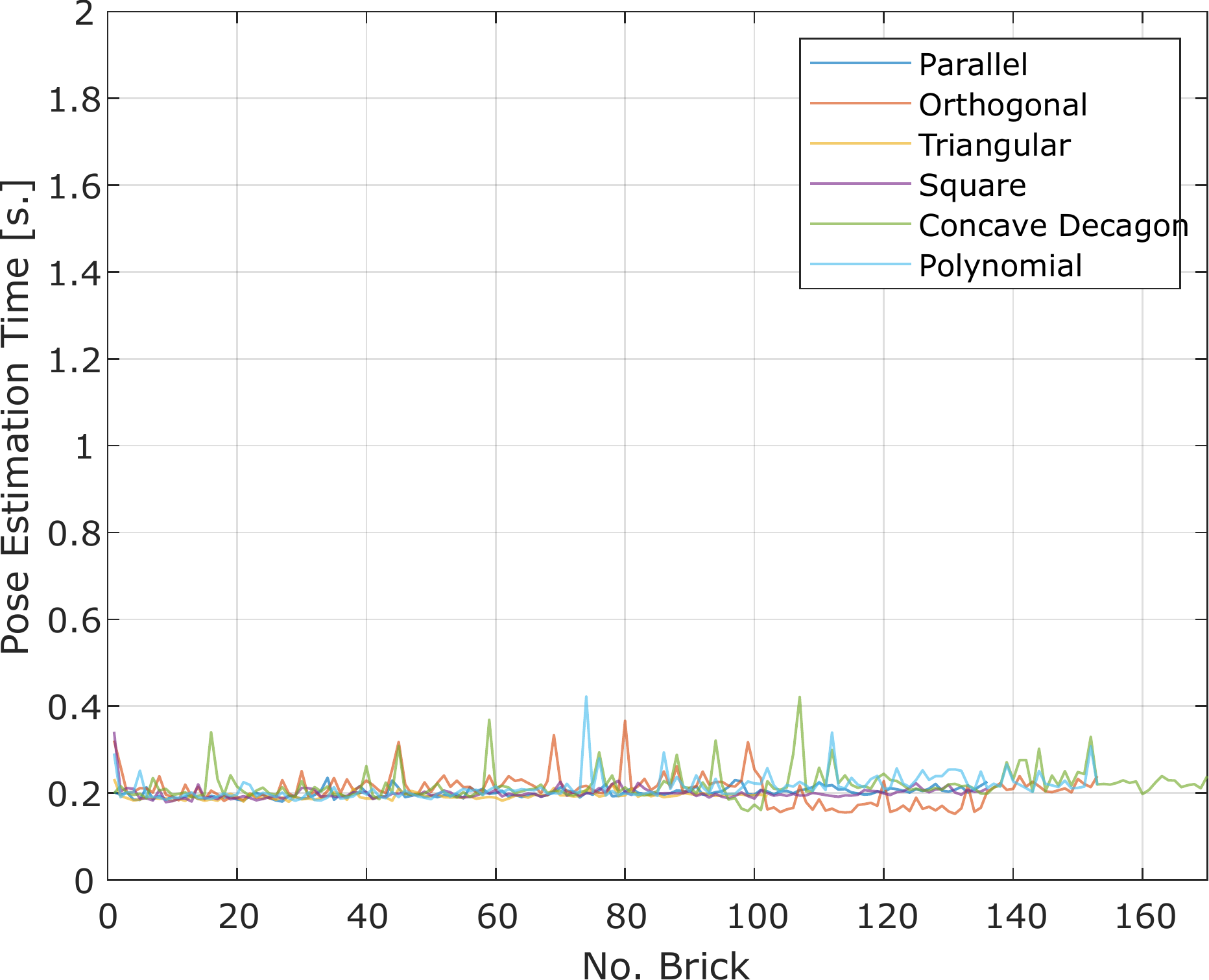}}
		\hfill
	\end{center}
	\caption{Performance metrics in the assembly of spiral brick columns.}
	\label{perfo}
\end{figure*}

\begin{figure*}[h]
\begin{center}
\includegraphics[width=0.98\textwidth]{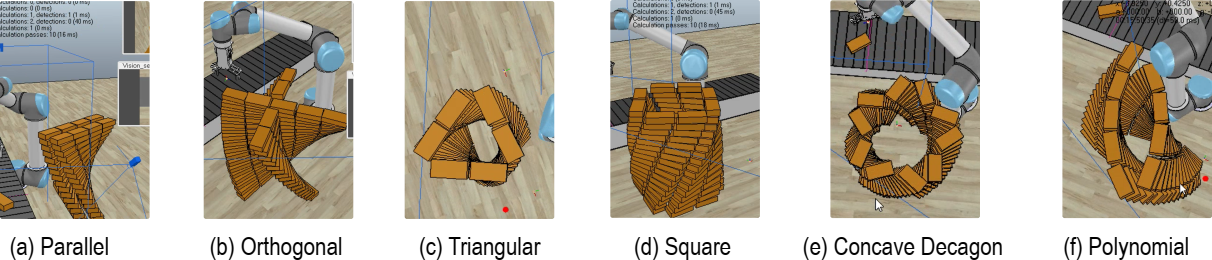}
\caption{Examples of the performance while assembling spiral brick columns.}
\label{simu}
\end{center}
\end{figure*}

\subsection{Pose Estimator}

To estimate the pose of incoming bricks from a conveyor, we use a depth camera close to the robot manipulator (see Fig. \ref{gen}), and compute the point cloud $\mathcal{P}$ satisfying the region of interest near the brick by using the pinhole cameral model and RANSAC with maximum likelihood and consensus (MLESAC) algorithm, in which errors are modeled by a mixture of Gaussians\cite{mlesac} instantaneously (Fig. \ref{pcloud}-(a-c)).



Furthermore, to estimate the pose of bricks, we find the minimum bounding box by using the \emph{calipper} algorithm with bounding box containing edges of the convex hull of the plane\cite{calipper83}. Fig. \ref{pcloud}-(d) shows an instance of pose estimation, in which the bounding box is in by blue color and the orientation vector is in red color.

\subsection{Task Execution}

The task planner has the main role of deciding the sequences of manipulator motion to enable the effective pick and place of bricks.

Given the pose $(\mathbf{p}, \alpha$) of the brick in the conveyor and its target pose $(\mathbf{p}_b, \alpha_b$), the manipulator proceeds as follows:

\begin{itemize}
  \item The gripper is aligned to the brick pose $(\mathbf{p}, \alpha$) by first aligning over the $x-y$ plane over the point $\mathbf{q}  = \mathbf{p} + \eta.\mathbf{\hat{k}}$ for $\eta > 0$. Then, the brick is picked up from the midpoints of its largest sides and is moved to $\mathbf{q}$.
  \item The planner finds a path that drives the robot from its configuration at $\mathbf{q}$ to the target pose $(\mathbf{p}_b, \alpha_b$) in the shortest path in Cartesian space (a straight line).
\item Move the gripper into the desired pose $(\mathbf{p}_b, \alpha_b$), open the gripper and move back to the position and wait for the gripper to grasp the next brick.
\item Continue and repeat all the above steps until all bricks are transported to their desired poses in the spiral brick columns.
\end{itemize}

\section{Computational Experiments}

In order to evaluate the effectiveness of our approach we used a simulation environment which represents the interactions and physics of rigid body dynamics. We used Matlab for 3D modeling and planning execution tasks and Coppelia Simulation environment for robot manipulation tasks. simulation parameters involve bricks with $w = 0.5$, $l = 0.1$, and height $h = 0.025$, $\lambda = 0.01$, $\kappa = 0.05$, $L = 17$, and $\eta = 1.25$. We used the geometries and configurations of base polygons as shown by Fig. \ref{models}. Fine tuning of the parameters is our of the scope of this paper.

In order to show the performance of our proposed approach, Fig. \ref{perfo} shows the errors in position and orientation in the spiral brick column, and the time used in pose estimation and robot pick and place. Also, Fig. \ref{simu} shows a number of examples of spiral brick columns in our simulation environment. As we can observe from Fig. \ref{perfo}-(a,b) the position error in terms of euclidean distance is less than o.025 with a number of spikes at certain bricks in orthogonal column, and the orientation error in terms of absolute terms coincides with a mean of about 0.025 degrees. There are spikes at certain bricks which give a maximum difference at 0.125 degrees.

Also, by observing Fig. \ref{perfo}-(c), the pose estimation is less than half a second with a mean of about 0.2 seconds, implying the real usefulness for real time interaction. Furthermore, by observing Fig. \ref{perfo}-(d), the trajectory time is the time needed for the robot assembly from pick to place in the desired position and orientation. Fig. \ref{perfo}-(d) sows that the concave decagon model takes larger time compared to other models such that maximum’s trajectory time per brick is 7 s.. Also, the orthogonal model takes the shortest time compared to other models such that the minimum time is 3 s.

\section{Final Notes}

In this paper, we have propose an approach for task-level planning of assembly of spiral brick columns. Our computational studies have shown the feasibility and efficiency of our approach to render spiral brick columns in simulation environments. Our study aims at elucidating the robot plans being effective for automated construction of aesthetic spiral brick columns.

\bibliography{mybiblio}
\bibliographystyle{IEEEtran}

\end{document}